\documentclass[10pt,twocolumn,letterpaper]{article}

\usepackage{iccv}
\usepackage{times}
\usepackage{epsfig}
\usepackage{graphicx}
\usepackage{amsmath}
\usepackage{amssymb}

\usepackage{subfigure}
\usepackage{algorithm}  
\usepackage{algorithmic} 

\usepackage{balance}

\graphicspath{{images/}}

\usepackage[pagebackref=true,breaklinks=true,letterpaper=true,colorlinks,bookmarks=false]{hyperref}

\iccvfinalcopy 

\def\iccvPaperID{1220} 

\ificcvfinal\fi
\begin{document}

\title{Deformable Distributed Multiple Detector Fusion for Multi-Person Tracking}

\author{Andy J Ma\\
Johns Hopkins University\\
{\tt\small andyjhma@cs.jhu.edu}
\and
Pong C Yuen\\
Hong Kong Baptist University\\
{\tt\small pcyuen@comp.hkbu.edu.hk}
\and
Suchi Saria\\
Johns Hopkins University\\
{\tt\small ssaria@cs.jhu.edu}
}


\maketitle

\begin{abstract}
   This paper addresses fully automated multi-person tracking in complex environments with challenging occlusion and extensive pose variations. Our solution combines multiple detectors for a set of different regions of interest (e.g., full-body and head) for multi-person tracking. The use of multiple detectors leads to fewer miss detections as it is able to exploit the complementary strengths of the individual detectors. While the number of false positives may increase with the increased number of bounding boxes detected from multiple detectors, we propose to group the detection outputs by bounding box location and depth information. For robustness to significant pose variations, deformable spatial relationship between detectors are learnt in our multi-person tracking system. On RGBD data from a live Intensive Care Unit (ICU), we show that the proposed method significantly improves multi-person tracking performance over state-of-the-art methods.
\end{abstract}

\vspace{-6pt}
\section{Introduction}

Monitoring human activities in complex environments are finding an increasing number of applications (e.g.,~\cite{chan2008review, corchado2008intelligent, Hwang2004}). Our current investigation is driven by the application of automated surveillance in a hospital, specifically, critical care units that house the sickest and most fragile patients. In 2012, the Institute of Medicine--the health arm of the National Academies of Sciences--released  their landmark report~\cite{smith2013best} on developing digital infrastructures that enable rapid learning health systems; one of their key postulates is the need for improvement technologies for measuring the care environment. Currently, simple measures such as whether the patient has moved in the last 24 hours, or whether the patient has gone unattended for several hours require manual observation by a nurse, which is highly impractical to scale.
Thus, this paper tackles fully-automated multi-person tracking, a first step needed towards automated surveillance in such complex indoor environments.

\begin{figure}
\centering
\includegraphics[width=0.9\linewidth]{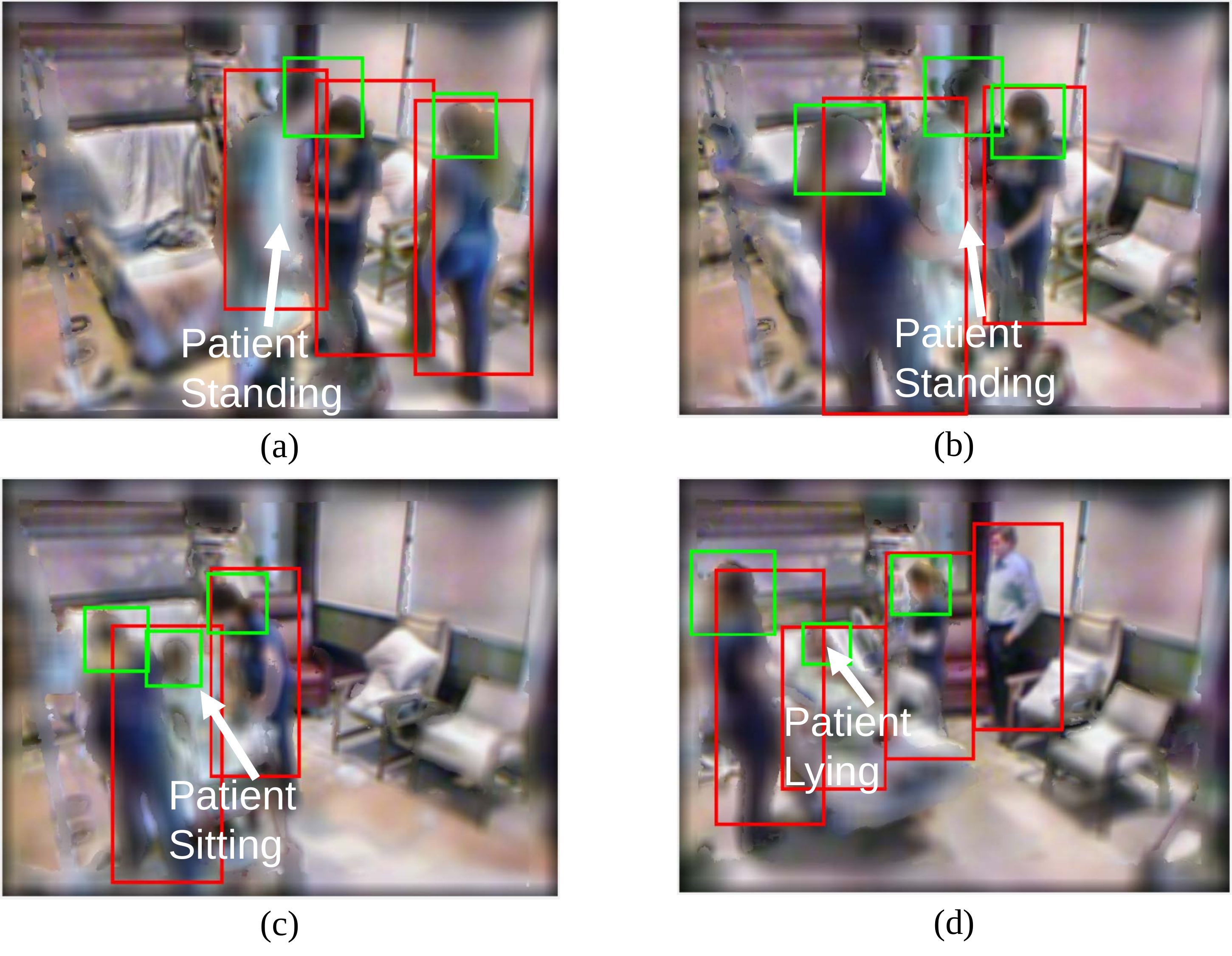}
\caption{Example images show detection results using full-body (Red) and head (Green) detectors trained by deformable part models~\cite{felzenszwalb2010object} (Better view in color). The head detector fails when (a) the two persons' heads get close to each other or (d) the head is far away from the camera and close to the wall (which makes the depth cue not discriminative). On the other hand, the full-body detector becomes less effective when two persons are close to each other as shown in (b) and (c). Moreover, human pose can change significantly, e.g., the patient goes from standing in (a) and (b) to sitting in (c) and lying in (d). (Due to privacy problem, color images captured in a real ICU room are blurred for de-identifying.)}
\vspace{-10pt}
\label{figExampleImgs}
\end{figure}


Person detection in images is an active research area~\cite{azizpour2012object, chen2014detect, dalal2005histograms, felzenszwalb2010object, xia2011human, zeng2010robust}. For example, a widely used detector is the one by Dalal and Triggs \cite{dalal2005histograms} that uses the histogram of oriented gradients to train a full-body detector. To tackle pose variations of articulated objects like humans, several works have introduced part-based models~\cite{azizpour2012object, chen2014detect, felzenszwalb2010object}. Deformable part models (DPMs) \cite{felzenszwalb2010object}, for example, train a discriminative detector with latent variables to jointly reason over individual parts and full-body locations. However, these do not take full advantage of the temporal dependence between successive images in video data.

To leverage temporal cues, visual tracking based approaches update the target location in a given frame conditioned on tracking results in the previous frame. Many appearance based tracking methods~\cite{babenko2009visual, grabner2006line, hare2011struck, lan2014multi, li2013survey, luonline, wu2013online, yao2013part, zhang2012real} have been developed to tackle challenges due to changes in illumination, occlusion and scale variations. Nevertheless, these methods assume a hand annotated region of interest is available at the onset, rendering these methods inapplicable in the fully automated setting.

To avoid manual annotation of the target for tracker initialization, tracking-by-detection techniques~\cite{andriluka2008people, andriyenko2010globally, andriyenko2012discrete, berclaz2011multiple, breitenstein2011online, chen2014multi, choi2013general, milan2013continuous, shu2012part, wen2014multiple, yang2012online} employ a single per-frame detector with temporal consistency constraints for tracking. In settings where occlusion is of concern, part-based models have been augmented with visibility variables corresponding to each part for occlusion reasoning~\cite{shu2012part, yang2012online}. Although these methods achieve convincing results for fully automated tracking, reliance on a single detector may be suboptimal as shown in Fig.~\ref{figExampleImgs}. To make use of the complementary information from multiple detectors, Choi \etal~\cite{choi2013general} propose to combine a set of detectors in a general multi-person tracking framework. However, their approach assumes that the spatial configuration between the detectors are fixed. Specifically, for a given detector output (e.g., the bounding box from a full-body detector),
their approach specifies a mask for where the other detector output (e.g., the bounding box from a face
detector) should lies. This does not scale to domains where individuals show significant pose variations (see examples in Fig.~\ref{figExampleImgs}).

To overcome these limitations, we propose to combine multiple detectors and model the deformable spatial relationship between them. More specifically, each detector seeks to detect a region of interest (e.g., full-body or head). On a single frame, a detector may yield one or more bounding boxes. Since bounding boxes from different detectors may belong to the same person, we first group the detection outputs to reduce the number of possible persons. Then, the deformable spatial relationship between different regions of interest are modeled by a least-square reconstruction constraint. At each frame, the inferred locations for persons are obtained by optimizing an objective function comprising of terms that penalize inferred person and individual detector locations that a) are distant from detector outputs, b) deviate from expected spatial structure, c) differ significantly in consecutive frames, and d) contain overlap in region-of-interest assignment for multiple persons (e.g. the inferred full-body locations for the two persons are at the same location). The proposed optimization objective is non-differentiable; we approximate each of these above-mentioned terms using a differentiable function, and develop a block gradient descent algorithm which provides an online estimate of the inferred person locations in the given video segment. To summarize, this paper makes the following key contributions:


$\bullet$ We develop a novel method for multi-person tracking by combining multiple detectors for a set of different regions of interest within a tracking-by-detection framework. Our method exploits the complementarity
of cues from multiple detectors, that is, targets missed under one detector may be detected by another detector. Since the proposed method combines detectors in the level of detection outputs (bounding boxes), it is flexible and can incorporate new state-of-the-art detectors as they become available (e.g., \cite{azizpour2012object, chen2014detect, dalal2005histograms, felzenszwalb2010object, xia2011human, zeng2010robust}). Moreover, these detectors can be geared towards different modalities (e.g., depth versus RGB versus RGBD).

$\bullet$ Our approach models the deformable spatial relationship between multiple detectors for multi-person tracking. We show that by doing so the tracking performance improves significantly for
tracking articulated objects like humans in domains with significant pose variations.

$\bullet$ Empirically, we show that multi-person tracking performance can be significantly improved by combining multiple detectors on a novel and challenging dataset obtained from an Intensive Care Unit (ICU).

\vspace{-2pt}
\section{Related Work}

Existing multi-target tracking methods can be divided into two categories: offline and online approaches.

The offline approach primarily used optimization based formulations in which the most likely trajectory is optimized over the entire video segment. In prior works, both discrete~\cite{andriyenko2010globally, berclaz2011multiple} and continuous~\cite{andriyenko2012discrete, milan2013continuous} formulations have been used for identifying latent person trajectories. Notably, in recent work by Milan~\etal~\cite{milan2013continuous}, they use temporal and appearance consistency based constraints coupled with an occlusion model and show state-of-the-art performance for multi-person tracking. While these methods process data in batch mode, they cannot be deployed to real-time system.

Online multi-target tracking methods~\cite{breitenstein2011online, liu2012automatic, okuma2004boosted} widely use particle filter based framework to associate the tracker with detection outputs. To make use of multiple complementary cues, Choi \etal~\cite{choi2013general} propose to combine multiple detectors to estimate the likelihood of the target observation.
They perform score-level fusion to combine the multiple detectors. To estimate
these scores, for a given detector output (e.g., the bounding box from a full-body detector), they specify a mask for where the other detector output (e.g., the bounding box from a face
detector) should lie. Scores are computed based on the overlap of a detector
output with this mask. In domains with significant pose variations, the relative positions of the different regions of interest (e.g., head versus the full-body) may vary greatly. Thus, we develop a more general-purpose framework
that does not require making this restrictive assumption.


\vspace{-2pt}
\section{Proposed Method}

Our method builds on the tracking-by-detection framework in~\cite{milan2013continuous}. Their approach uses a single detector to do individual image level detection. They formulate an energy functional comprising spatial consistency and temporal consistency terms related to appearance and position. We extend this work to the multiple detector setting. In order to tackle pose variations, we model the relationship between detectors within a single frame using a deformable spatial model. Furthermore, our approach tackles tracking in an online setting.


We begin by giving an overview of our proposed method. We use a collection of existing detection methods to train $K$ detectors; each detector is geared towards detecting a region of interest (e.g., full-body or head). These detectors need not be semantically distinct, that is, one may choose to use more than one detector for a region of interest. We assume that ground truth bounding boxes corresponding to the regions of interest for each of the detectors are available in the training data. Let the number of regions of interest be $L$. We maintain $M^t$, the number of persons in any given frame $t$ as an unknown. Since bounding boxes from the $K$ detectors may come from the same person, we first group the detection outputs into $N^t$ groups as elaborated in Section~\ref{secGroup}. In each group, there is at least one bounding box to represent all the detection evidence for one person. Deformable spatial relationship is modeled by minimizing the least-square errors between the expected location and projected one as introduced in Section~\ref{secDeform}. The first frame uses an initialization step described in Section~\ref{secInit} to infer the number of persons and their inferred bounding box locations. Thereafter, to determine the detector bounding box locations at time $t$, we proceed using three steps. First, a preprocessing step is used to determine for each person in the previous frame, whether they exist in the current frame. Then, an optimization objective is formulated for optimizing the bounding box locations at time $t$ conditioned on the inferred bounding box locations at time $t-1$ and the detector outputs at time $t$. Finally, a postprocessing step is used to update the appearance model and assign identities for each of the inferred bounding box locations. These steps are discussed in Section~\ref{secUpdate}.

Notations: We denote output bounding boxes from the $k$-th trained detector in the $t$-th frame as $D^t_k(1), \cdots, D^t_k(N^t_k)$ and corresponding detection scores normalized by a exponential function as $w^t_k(1), \cdots, w^t_k(N^t_k)$, where $N^t_k$ is the number of detected bounding boxes. Let the number of persons in a frame be $M^t$. For the $m$-th person, the corresponding bounding boxes from $L$ regions of interest are defined by $X^t_1(m), \cdots, X^t_L(m)$. Let $X^t$ to refer to the collection of inferred locations of regions of interest for all persons in the frame. In this paper, each bounding box is represented by coordinates of the upper-left and lower-right corners.

\vspace{-2pt}
\subsection{Grouping Bounding Boxes from Multiple Detectors}\label{secGroup}


Let us consider two bounding boxes $D^t_k(n)$ and $D^t_{k'}(n')$ from any two detectors $k$ and $k'$, respectively. We calculate the probability that these two bounding boxes are from the same person as,
\begin{equation}\label{eqProbAll}
\begin{split}
    p = a p_{\textrm{over}} + (1-a) p_{\textrm{depth}}
\end{split}
\end{equation}
where $a$ is a positive weight, $p_{\textrm{over}}$ and $p_{\textrm{depth}}$ measure the overlapping ratio and depth similarity between two bounding boxes, respectively. These two probability scores are
\begin{equation}\label{eqProbOver}
\begin{split}
    p_{\textrm{over}} = \frac{\left|D^t_k(n) \cap D^t_{k'}(n')\right|}{\min \left( |D^t_k(n)|, |D^t_{k'}(n')| \right)}
\end{split}
\end{equation}
\begin{equation}\label{eqProbDepth}
\begin{split}
    & p_{\textrm{depth}} = \frac{1}{2} e^{\frac{-(v_k^t(n)-v_{k'}^t(n'))^2}{2 \sigma_k^t(n)^2}} + \frac{1}{2} e^{\frac{-(v_k^t(n)-v_{k'}^t(n'))^2}{2 v_{k'}^t(n')^2}}
\end{split}
\end{equation}
where $v$ and $\sigma$ denote the mean and standard deviation of the depth values inside bounding box, respectively.

\begin{figure}
\centering
\subfigure[Color image]{\label{figGroupImg1}\includegraphics[width=0.4\linewidth]{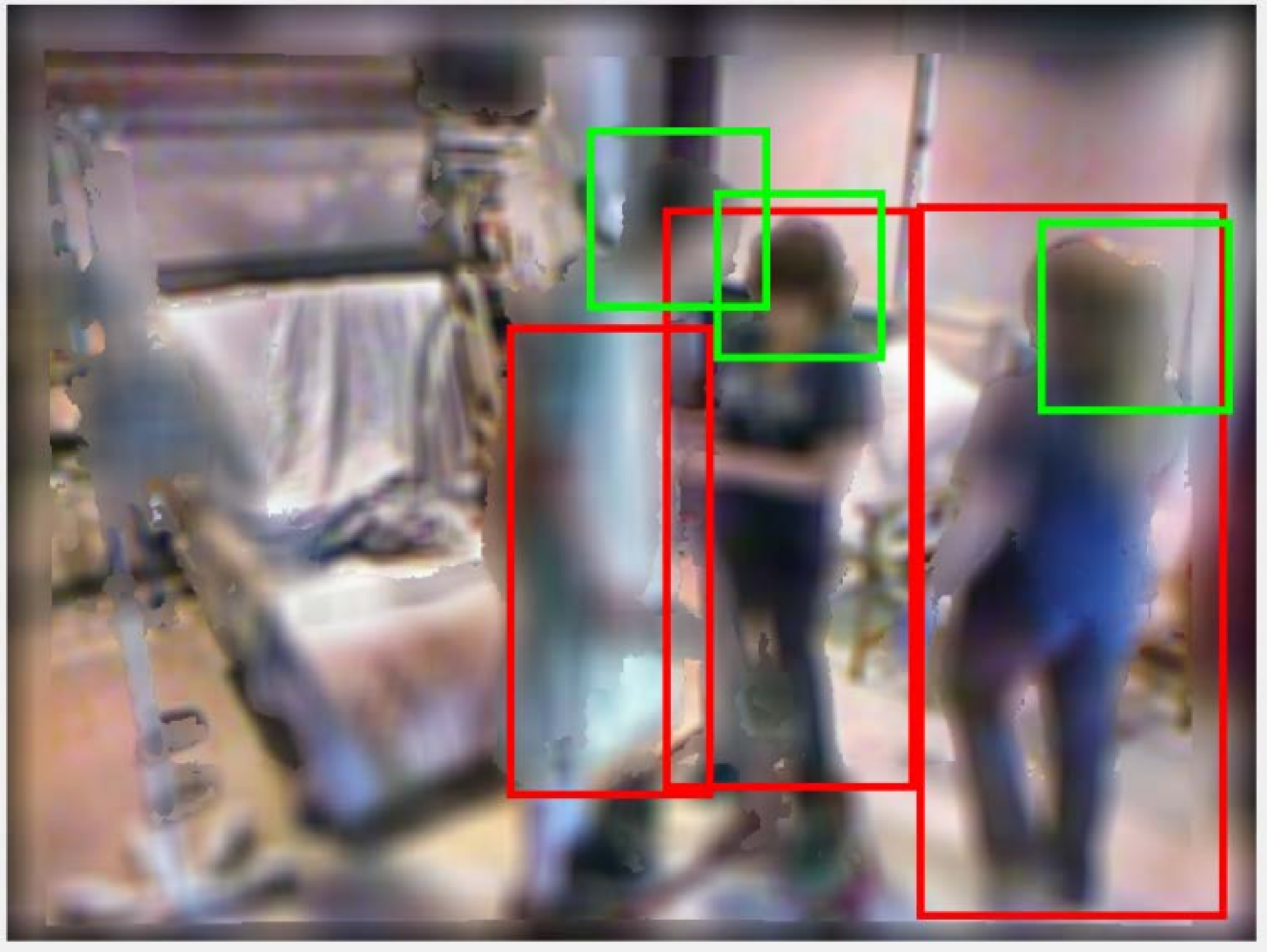}} \hspace{4pt}
\subfigure[Depth image]{\label{figGroupImg2}\includegraphics[width=0.4\linewidth]{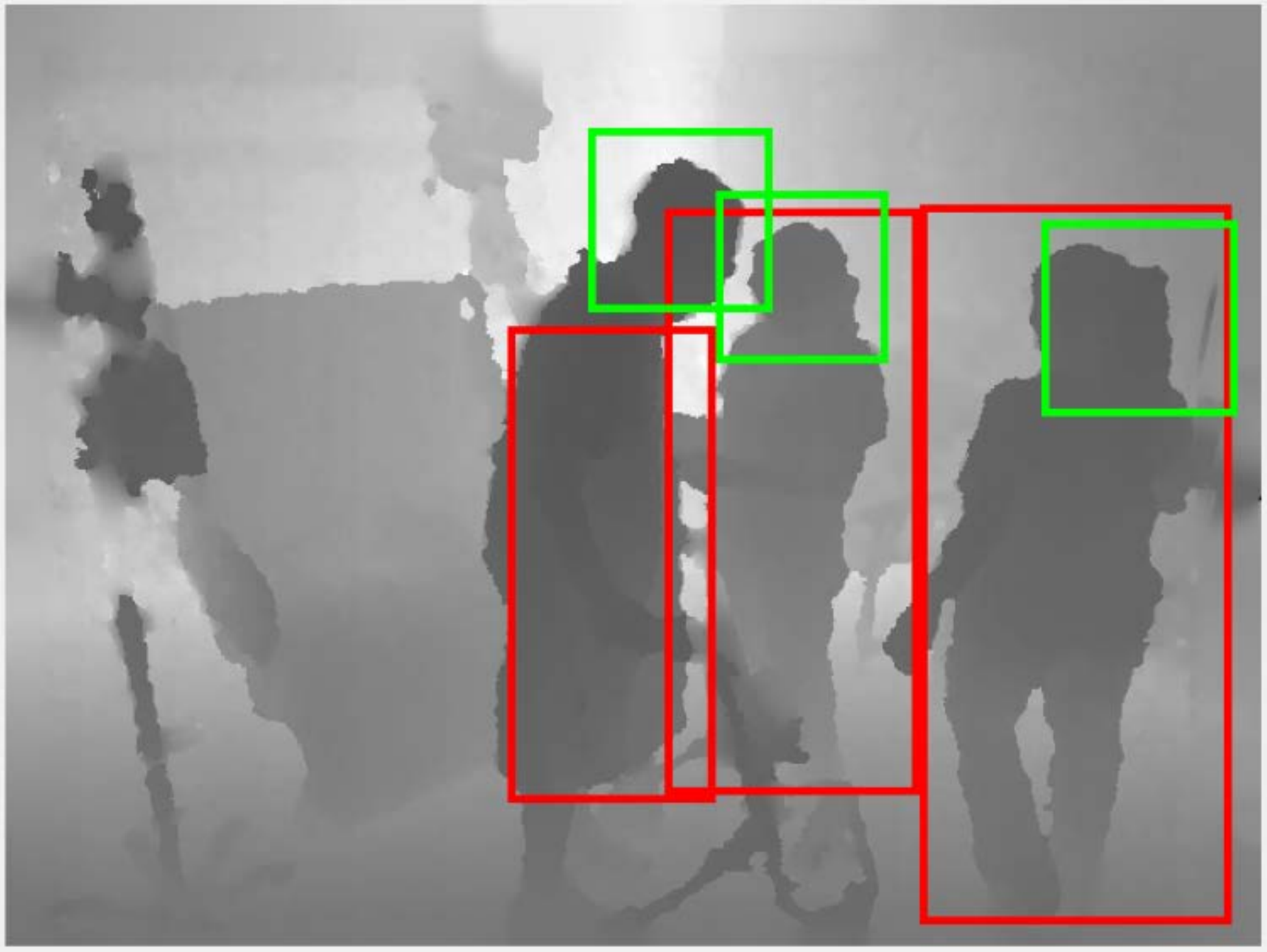}}
\caption{Example images to demonstrate the motivation for bounding boxes grouping (better view in color)}
\vspace{-10pt}
\label{figGroup}
\end{figure}

If two bounding boxes are overlapped with each other, e.g., the head is contained in the full-body, such probability should be large. Thus, we define the probability using the first term~\eqref{eqProbOver} in~\eqref{eqProbAll}. However, as shown in Fig.~\ref{figGroup}, it is possible that $D^t_k(n)$ (e.g., the full-body bounding box on the left) does not overlap with any bounding boxes from detector $k'$ (e.g., head). This motivates another term for the grouping probability. Since regions of interest from the same person should have similar depth values, the depth similarity scores can help to better group the bounding boxes. Thus, the second term is defined by~\eqref{eqProbDepth} based on the depth information.

To group the bounding boxes from any two detectors $k$ and $k'$, we compute the grouping probability for each pair $n$ and $n'$ by~\eqref{eqProbAll} to get a similarity matrix $P^t_{kk'}$. The largest element in $P^t_{kk'}$ is selected iteratively. Since one bounding box from detector $k$ is corresponding to only one bounding box from detector $k'$, the elements in the $n$-th row of $P^t_{kk'}$ are removed for the selected bounding box pair $(n,n')$. Similarly, we remove the $n'$-th column of $P^t_{kk'}$. If the selected elements are lager than a threshold $\tau$, the bounding pair $(n,n')$ is considered to come from the same person.

To extend the algorithm to any number of detectors $K$, we first generate groups of bounding boxes $B^{t}_2(1), \cdots, B^t_2(N^{t}_2)$ for two detectors by the grouping procedures mentioned above. Similarly, $B^{t}_2(1), \cdots, B^t_2(N^{t}_2)$ and bounding boxes from another detector can be grouped to obtain $B^{t}_3(1), \cdots, B^t_3(N^{t}_3)$. After $K-1$ iterations, we get $N^t$ groups of bounding boxes $B^{t}(1), \cdots, B^t(N^{t})$ to represent $N^t$ possible persons.

With the groups of bounding boxes, we assume that the detector outputs $D^t_k(n)$ are more likely to be spatially contiguous with true positions for those regions of interest in the image. In other words, e.g. the head is more likely to be located near the candidate locations suggested by the head detector output. Thus, the cost function for this detection matching constraint is devised as
\begin{equation}\label{eqDetCost}
\begin{split}
    & E_{det}(X^{t},M^{t}) = \\
    & \sum_{n'=1}^{N^t} \min_{1\leq m \leq M^t} \sum_{D^t_k(n) \in B^t(n')} w^t_k(n) \left\|D^t_k(n) - X^t_{\iota(k)}(m)\right\|^2
\end{split}
\end{equation}
where $\iota$ maps the $k$-th detector to the $l$-th region of interest. If the detection confidence $w^t_k(n)$ is large, we assume greater likelihood of having a tracked bounding box $X^t_k(m)$ close to $D^t_k(n)$. Therefore, we use $w^t_k(n)$ to define the weighted penalty for each detected bounding box $D^t_k(n)$ in \eqref{eqDetCost}. Note that, as presented in \eqref{eqDetCost}, this term is not differentiable. To ensure differentiability of the cost function, we approximate the minimum operation by
\begin{equation}\label{eqMinApprox}
\begin{split}
    \min_{1\leq m \leq M} \{z_m\} & = \lim_{\alpha\rightarrow-\infty} S_\alpha(\{z_m\})  \\
    S_\alpha(\{z_m\}) & = \frac{\sum_{m=1}^{M} z_m e^{\alpha z_m}}{\sum_{m=1}^{M} e^{\alpha z_m}}
\end{split}
\end{equation}
With the approximation by \eqref{eqMinApprox}, the cost function becomes
\begin{equation}\label{eqDetCostApprox}
\begin{split}
    & E_{det}(X^{t},M^{t}) = \\
    & \sum_{n'=1}^{N^t} S_\alpha \left( \left\{ \sum_{D^t_k(n) \in B^t(n')} w^t_k(n) \left\|D^t_k(n) - X^t_{\iota(k)}(m)\right\|^2 \right\} \right)
\end{split}
\end{equation}

\subsection{Modeling Deformable Spatial Relationship}\label{secDeform}
In addition to matching candidate detections, we wish to take advantage of the expected spatial structure between regions of interest for any given articulated pose variations. For example, when standing, we expect the location of the head bounding box to be in the top region of the full-body bounding box. Moreover, for two detectors that seek to detect the same region of interest (e.g., if we employed two different head detectors), the expected bounding box coordinates would be expected to be in the same location. We model these deformable spatial correlations using a function that penalizes deviations from expected configurations.

Let $A_{ll'}^c$ be a projection matrix which learns the spatial relationship between the bounding box locations of two regions of interest $l$ and $l'$ for a given pose $c$. Specifically, given training data, for every pose $c \in \{1, \cdots, C\}$, it learns a mapping from the expected location $X^{t}_l(m)$ from the $l$-th region to the location $X^{t}_{l'}(m)$ from the $l'$-th one. Now, for any two proposed bounding boxes $X^{t}_{l'}(m)$ and $X^{t}_{l}(m)$ at frame $t$ for individual $m$, the deviation from the expected spatial configuration is quantified as the error between the expected location of the bounding box for the second region of interest (e.g., head) conditioned on the location of the first region of interest (e.g., full-body). The total cost is computed by summing for each of the $M^t$ individuals, the minimum cost for each of the $C$ subcategories, i.e.,
\begin{equation}\label{eqSpaCost}
\begin{split}
    E_{spa}(X^{t}, M^{t}) = \sum_{m=1}^{M^t} \min_{1\leq c \leq C} \sum_{l \neq l'}\|A_{ll'}^c X^{t}_l(m) - X^t_{l'}(m)\|^2
\end{split}
\end{equation}
Again, due to the minimization, this term is not differentiable. Similar to the relaxation we employ for the detection matching constraint, we approximate the minimization operation by \eqref{eqMinApprox}. The resulting cost function \eqref{eqSpaCost} becomes
\begin{equation}\label{eqSpaCostApprox}
\begin{split}
    & E_{spa}(X^{t}, M^{t}) \\
    & \quad = \sum_{m=1}^{M^t} S_\alpha \left( \left\{ \sum_{l \neq l'}\|A_{ll'}^c X^{t}_l(m) - X^{t}_{l'}(m)\|^2 \right\} \right)
\end{split}
\end{equation}

To obtain the projection matrices $A_{ll'}^1, \cdots, A_{ll'}^C$, we employ the spatial features in~\cite{chen2014detect} to separate the normalized training data into $C$ subcategories by \emph{k}-means clustering. Then, $A_{ll'}^c$ is trained by solving the regularized regression problem with the training data in the $c$-th subcategory.

\subsection{Optimized Tracker Initialization}\label{secInit}
To determine the optimal tracker initialization at time $t=1$, we aggregate the detection matching~\eqref{eqDetCostApprox} and deformable spatial~\eqref{eqSpaCostApprox} cost functions with two additional constraints (akin to~\cite{milan2013continuous} for single detector) considering the mutual exclusion and regularization for multiple persons, i.e.,
\begin{equation}\label{eqExcCost}
\begin{split}
    E_{exc}(X^{t}, M^{t}) = \sum_{l=1}^L \sum_{m \neq m'}  \frac{1}{\|X^t_l(m) - X^t_l(m')\|^2}
\end{split}
\end{equation}
\begin{equation}\label{eqRegCost}
\begin{split}
    E_{reg}(X^{t}) = M^t
\end{split}
\end{equation}
By linearly weighting the cost functions \eqref{eqDetCostApprox} \eqref{eqSpaCostApprox} \eqref{eqExcCost} \eqref{eqRegCost}, the joint optimization is given by
\begin{equation}\label{eqOptSpatial}
\begin{split}
    \min_{X^{t}, M^{t}} \lambda_{det} E_{det} + \lambda_{spa} E_{spa} + \lambda_{exc} E_{exc} + \lambda_{reg} E_{reg}
\end{split}
\end{equation}
We solve \eqref{eqOptSpatial} incrementally for increasing values of $M^t$. Given candidate initializations, \eqref{eqOptSpatial} can be optimized using gradient descent. Since \eqref{eqOptSpatial} is not convex, clever initialization can both improve the computational efficiency and quality of the solution.

We can leverage the deformable spatial model to yield useful initializations. For each of the detected bounding boxes $D^t_k(n)$ for the $k$-th detector, for the $c$-th subcategory, using the projection matrices $A_{\iota(k)1}^c, \cdots, A_{\iota(k)L}^c$, we predict the expected locations for each of the bounding boxes of other regions of interest as
\begin{equation}\label{eqConfigEst}
\begin{split}
     \hat{X}^{tc}_1(n^t_k), \cdots, \hat{X}^{tc}_L(n^t_k) = (A_{\iota(k)1}^c, \cdots, A_{\iota(k)L}^c) D^t_k(n)
\end{split}
\end{equation}
These predictions yield $C \times \sum_{k=1}^K N^t_k$ candidate sets of bounding boxes for persons in the image, where each set corresponds to all the regions of interest for pose $c$. This approach of predicting bounding box locations for the other detectors conditioned on the location for a given detector is useful when the original detector fails to identify any candidate locations. For example, as shown earlier in Fig.~\ref{figExampleImgs}(a)(d), while the head detector cannot successfully detect the head of some persons, prediction using a full-body detector may successfully localize the head bounding box. For $M^t=1$, we select the candidate set of bounding boxs that minimizes \eqref{eqOptSpatial}. As $M^t$ is increased, we retain the previous initializations and augment the set by selecting the next best candidate bounding box locations using the same criterion.

\renewcommand{\algorithmicrequire}{\textbf{Input:}}
\renewcommand{\algorithmicensure}{\textbf{Output:}}
\begin{algorithm}[t]\small
\caption{Initialization at the first frame $t=1$}
\label{alg:Init}
\begin{algorithmic}[1]
\REQUIRE Detected bounding boxes $D^t_1(1), \cdots, D^t_K(N^t_K)$ and corresponding weights $w^t_1(1), \cdots, w^t_K(N^t_K)$;
\STATE For each $D^t_k(n)$ generate $C$ sets of bounding boxes by \eqref{eqConfigEst} for $C$ types of deformations;
\STATE Set the objective cost as infinite;
\STATE Let $M^t=1$;
\STATE Select a set $\hat{X}^{tc}_1(n^t_k), \cdots, \hat{X}^{tc}_L(n^t_k)$ that minimizes \eqref{eqOptSpatial} as the initialized solution;
\STATE Employ gradient descend to obtain the optimal solution $X^t_1(1), \cdots, X^t_L(M^t)$ for fixed $M^t$;
\IF{objective cost decreases}
    \STATE Save $X^t_1(1), \cdots, X^t_L(M^t)$ as the optimal solution;
    \STATE Let $M^t = M^t + 1$;
    \STATE Add $\hat{X}^{tc}_1(n^t_k), \cdots, \hat{X}^{tc}_L(n^t_k)$ to the previous initialization sets by minimizing \eqref{eqOptSpatial};
    \STATE Go to step 5;
\ELSE
    \RETURN the saved optimal solution;
\ENDIF
\ENSURE Optimal bounding boxes $X^t_1(1), \cdots, X^t_L(M^t)$.
\end{algorithmic}
\end{algorithm}

In summary, the algorithmic procedures for tracker initialization at the first frame $t=1$ is given in Algorithm~\ref{alg:Init}.

\subsection{Online Tracking Update}\label{secUpdate}
In the $t$-th frame, we are not only given the detection outputs $D^t_1(1), \cdots, D^t_K(N^t_K)$ and corresponding weights $w^t_1(1), \cdots, w^t_K(N^t_K)$ but also the tracked bounding boxes $X^{t-1}_1(1), \cdots, X^{t-1}_L(M^{t-1})$ in the previous frame.

\textbf{Preprocessing:} Before introducing the cost functions to ensure the temporal consistence, the inferred bounding boxes in the previous frame are preprocessed to remove the persons who are likely to be out of the camera in the current frame. The velocity $v_l^{t-1}(m)$ of the position of the $l$-th region of interest for person $m$ at time $t-1$ is calculated by
\begin{equation}\label{eqVel}
\begin{split}
    v_l^{t-1}(m) = X^{t-1}_l(m) - X^{t-2}_l(m)
\end{split}
\end{equation}
With the velocity $v_l^{t-1}(m)$, the current bounding box position $\tilde{X}^t_l(m)$ can be estimated by
\begin{equation}\label{eqPred}
\begin{split}
    \tilde{X}^t_l(m) = X^{t-1}_l(m) + v_l^{t-1}(m)
\end{split}
\end{equation}
If the estimated position $\tilde{X}^t_l(m)$ is close to the image boundary and there is no detected bounding boxes around $\tilde{X}^t_l(m)$ for $l = 1, \cdots, L$, person $m$ is thought to be out of the camera in the current frame. The temporal consistence is only valid for the $M^{t-1}_*$ persons who are still in the view of the camera at time $t$.

\textbf{Optimization:} For trajectory consistence, we define the cost function as
\begin{equation}
\begin{split}
    E_{tra}(X^t) = \sum_{l=1}^L \sum_{m=1}^{M^{t-1}_*} \left\|v_l^{t}(m) - v_l^{t-1}(m)\right\|^2
\end{split}
\end{equation}
where velocity is defined as in \eqref{eqVel}. In addition to motion based consistency, we also impose appearance based consistency. This implements the natural assumption that a person between two consecutive frames will likely look similar. Let $\psi$ be a normalized appearance feature extraction function. If the appearances of two bounding boxes $X^{t-1}_l(m)$ and $X^{t}_l(m)$ are similar, the inner product of their appearance feature vectors is large. Thus, we define the cost function for the appearance consistence term as
\begin{equation}
\begin{split}
    E_{app}(X^t) = - \sum_{l=1}^L \sum_{m=1}^{M^{t-1}_*} \psi\left(X^{t-1}_l(m)\right) \cdot \psi\left(X^{t}_l(m)\right)
\end{split}
\end{equation}

To ensure differentiability, we follow~\cite{milan2013continuous} to define the feature extraction function $\psi$ as
\begin{equation}
\begin{split}
    \psi_b\left(X^{t}_l(m)\right) & = \sum_{\textbf{\emph{p}}} H_b(\textbf{\emph{p}}) \mathcal{N}\left(\textbf{\emph{p}}; \mu_l^t(m), \Sigma_l^t(m)\right)\\
\end{split}
\end{equation}
where $\psi_b$ is the $b$-th element of the feature vector $\psi$, $\mathcal{N}(.)$ is a Gaussian distribution with the mean and covariance matrix of $\mu_l^t(m)$ and $\Sigma_l^t(m)$, respectively. $H_b(\textbf{\emph{p}})$ is a binning function of the $b$-th bin for the image.

Since the upper-left and lower-right corners are used to represent the inferred bounding boxes $X^{t}_l(m)$, the mean $\mu_l^t(m)$ is computed by
\begin{equation}
\begin{split}
    \mu_l^t(m) & = \left(
                     \begin{array}{c}
                       \emph{\textbf{a}}_1 \cdot X^{t}_l(m) \\
                       \emph{\textbf{a}}_2 \cdot X^{t}_l(m) \\
                     \end{array}
                   \right) \\
\end{split}
\end{equation}
where $\cdot$ denotes the inner product, $\emph{\textbf{a}}_1$ and $\emph{\textbf{a}}_2$ are vectors for mean computation, i.e.  $\emph{\textbf{a}}_1 = (\frac{1}{2}, 0, \frac{1}{2}, 0)^T, \emph{\textbf{a}}_2 = (0, \frac{1}{2}, 0, \frac{1}{2})^T$. On the other hand, we assume that the covariance matrix $\Sigma_l^t(m)$ is diagonal. Under normal distribution, 95.45\% of the values lie within two standard deviations of the mean. Thus, we define the standard deviations as half of size of the bounding box so that 95.45\% of energy concentrates with the bounding box, i.e.
\begin{equation}
\begin{split}
    \Sigma_l^t(m) & = \left(
                      \begin{array}{cc}
                        \frac{\left(\textbf{\emph{b}}_1 \cdot X^{t}_l(m)\right)^2}{2^2} & 0 \\
                        0 & \frac{\left(\textbf{\emph{b}}_2 \cdot X^{t}_l(m)\right)^2}{2^2} \\
                      \end{array}
                    \right)
\end{split}
\end{equation}
where $\emph{\textbf{b}}_1 = (-1, 0, 1, 0)^T$ and $\emph{\textbf{b}}_2 = (0, -1, 0, 1)^T$

Combining the trajectory and appearance consistence constraints with the cost functions derived in Section \ref{secInit}, the tracked bounding boxes are updated by solving the following optimization problem,
\begin{equation}\label{eqOptSpaTem}
\begin{split}
    \min_{X^t, M^t} & \lambda_{det} E_{det} + \lambda_{spa} E_{spa} + \lambda_{exc} E_{exc} \\
    & \qquad + \lambda_{tra} E_{tra} + \lambda_{app} E_{app} + \lambda_{reg} E_{reg}
\end{split}
\end{equation}
This optimization problem can be solved incrementally by increasing values of $M^t$ similar to procedure used in Section~\ref{secInit}. Differently, we have at least $M^{t-1}_*$ persons appear in the frame $t$, so the initial value of $M^t$ is set as $M^{t-1}_*$. The bounding box locations of these $M^{t-1}_*$ persons are initialized as in \eqref{eqPred}. As $M^t$ is increased, we retain the previous initializations and augment the set by selecting the next best candidate bounding box location akin to the procedure used in Section~\ref{secInit}. More specifically, as before, we obtain the set of candidate bounding boxes as in \eqref{eqConfigEst}. We then initialize using the set that minimizes \eqref{eqOptSpaTem}.


\renewcommand{\algorithmicrequire}{\textbf{Input:}}
\renewcommand{\algorithmicensure}{\textbf{Output:}}
\begin{algorithm}[t]
\caption{Temporal updating at time $t$}
\label{alg:Update}
\begin{algorithmic}[1]\small
\REQUIRE Detected bounding boxes $D^t_1(1), \cdots, D^t_K(N^t_K)$, corresponding weights as $w^t_1(1), \cdots, w^t_K(N^t_K)$, previous tracking results $X^{t-1}_1(1), \cdots, X^{t-1}_L(M^{t-1})$ and appearance models of all the previous targets $\phi^{t-1}_1(1), \cdots, \phi^{t-1}_L(M^{t-1}_{all})$;
\STATE For each $D^t_k(n)$ generate $C$ sets of bounding boxes by \eqref{eqConfigEst} for $C$ types of deformations;
\STATE Preprocess $X^{t-1}_1(m), \cdots, X^{t-1}_L(m)$;
\STATE Set the objective cost as infinite;
\STATE Let $M^t = M^{t-1}_*$;
\STATE Initialize the solution as $\tilde{X}^t_1(1), \cdots, \tilde{X}^t_L(M^{t-1}_*)$;
\STATE Employ gradient descend to obtain the optimal solution $X^t_1(1), \cdots, X^t_L(M^t)$ for fixed $M^t$;
\IF{objective cost decreases}
    \STATE Save $X^t_1(1), \cdots, X^t_L(M^t)$ as the optimal solution;
    \STATE Let $M^t = M^t + 1$;
    \STATE Add $\hat{X}^{1c}_1(n^1_k), \cdots, \hat{X}^{1c}_L(n^1_k)$ to the previous initialization sets by minimizing \eqref{eqOptSpaTem};
    \STATE Go to step 6;
\ENDIF
\STATE Assign identity to current targets $m$ ($1 \leq m \leq M^{t-1}_*$) as the same as the previous ones;
\STATE Calculate target matching confidence $\eta(p')$ for $p' > M^{t-1}_*$ by~\eqref{eqMatchCond};
\IF{$\eta(p') < \delta$}
    \STATE Assign a new identity to the current person $p'$;
\ELSE
    \STATE Assign the identity of the current person $p'$ as $p$ that achieves $\eta(p')$;
\ENDIF
\STATE Update appearance models by~\eqref{eqAppUp};
\ENSURE Optimal bounding boxes $X^t_1(1), \cdots, X^t_L(M^t)$, updated appearance models $\phi^{t}_1(1), \cdots, \phi^{t}_L(M^{t}_{all})$  and target identities.
\end{algorithmic}
\end{algorithm}

\textbf{Postprocessing:} After locating the optimal positions of persons in the current frame, identities of them must be assigned. For the persons which appear in the previous frame, their identities are the same as that in the previous frame. For a person $p'$ detected in this frame that does not have a corresponding detection in the previous frame, we use the appearance model to determine whether this detection corresponds to an individual previously seen, or a new individual. Towards this, we maintain and incrementally update $\Phi^{t-1}$, the appearance model for all individuals previously seen. Assume there are $M^{t-1}_{all}$ total number of individuals observed by time $t-1$. For the $p$-th individual and the $l$-th region of interest, the appearance at $t-1$ is denoted by $\phi^{t-1}_l(p)$. We determine the maximum similarity score $\eta(p')$ between the appearance of $p'$ and all individuals in $\Phi^{t-1}$ as
\begin{equation}\label{eqMatchCond}
\begin{split}
    \eta(p') = \max_{1 \leq p \leq M^{t-1}_{all}} \max_{1 \leq l \leq l} \phi^{t-1}_l(p) \cdot \psi\left(X^{t}_l(p')\right)
\end{split}
\end{equation}
If $\eta(p')$ is larger than a positive threshold $\delta$, the identity is assigned to $p$ that achieves $\eta(p')$. Otherwise, we add $p'$ to $\Phi^{t}$. Additionally, we update appearance models in $\Phi^{t}$ for individuals that appear in frame $t$ as follows:
\begin{equation}\label{eqAppUp}
\begin{split}
    \phi^{t}_l(p') = \frac{F^{t-1}(p') \phi^{t-1}_l(p') + \psi\left(X^{t}_l(p')\right)}{F^{t-1}(p') + 1}
\end{split}
\end{equation}
where $F^{t-1}(p')$ is the number of frames containing the $p'$-th person. The complete procedure for the online updating step is detailed in Algorithm~\ref{alg:Update}.

\begin{table*}[htbp]
  \centering\small
    \begin{tabular}{|r|r||r|r||r|r|r||r|r|}
    \hline
    Tracker & full-body Detector & MOTA  $\uparrow$ & MOTP $\uparrow$ &  FP $\downarrow$  &  FN $\downarrow$  & IDs $\downarrow$ & Recall  $\uparrow$ & Precision $\uparrow$ \\
    \hline
    Choi~\etal~\cite{choi2013general} & Pre-Trained   & -18.88 & 66.98& 32.10 & 78.68 & {8.10} & 21.32 & 38.63 \\
    \hline
    Choi~\etal~\cite{choi2013general} & ICU 1-Component DPM & -45.12 & 56.54& 57.36 & 76.82 & 10.94 & 23.18 & 27.06 \\
    Milan~\etal~\cite{milan2013continuous} & ICU 1-Component DPM & -38.61 & 62.79& 55.76 & 56.92 & 25.94 & 43.08 & 43.39 \\
    Ours & ICU 1-Component DPM & 13.50 & 67.04& 18.11 & 48.57 & 19.82 & 51.43 & 74.13 \\
    \hline
    Milan~\etal~\cite{milan2013continuous} & ICU 3-Component DPM & -15.21 & 68.03& 45.17 & 37.40 & 32.64 & 62.60 & 61.25 \\
    Ours & ICU 3-Component DPM & \textbf{29.14} & \textbf{68.64}& {12.77} & {36.43} & 21.66 & {63.57} & {83.06} \\
    \hline
    \end{tabular}%
  \caption{Results (\%) on ICU data ($\uparrow$ and $\downarrow$ mean higher and lower scores for better performance, respectively)}
  \vspace{-4pt}
  \label{tabQuanICU}%
\end{table*}%

\begin{table*}[htbp]
  \centering\small
    \begin{tabular}{|r|r||r|r||r|r|r||r|r|}
    \hline
    Tracker & full-body Detector & MOTA  $\uparrow$ & MOTP $\uparrow$ &  FP $\downarrow$  &  FN $\downarrow$  & IDs $\downarrow$ & Recall  $\uparrow$ & Precision $\uparrow$ \\
    \hline
    Choi~\etal~\cite{choi2013general} & Pre-Trained  & 20.20 & 57.62& 20.90 & 57.62 & 1.28 & 42.38 & 67.67 \\
    \hline
    Choi~\etal~\cite{choi2013general} & ICU 1-Component DPM & 16.28 & 59.83& {6.72}  & 76.43 & {0.56} & 23.57 & {80.21} \\
    Milan~\etal~\cite{milan2013continuous} & ICU 1-Component DPM & 21.68 & \textbf{74.61} & 34.10 & {40.27} & 3.95  & {59.73} & 64.37 \\
    Ours & ICU 1-Component DPM & \textbf{26.91} & 70.37 & 13.91 & 57.07 & 2.10  & 42.93 & 74.97 \\
    \hline
    \end{tabular}%
  \caption{Results (\%) on RGBD Pedestrian Dataset ($\uparrow$ and $\downarrow$ mean higher and lower scores for better performance, respectively)}
  \vspace{-6pt}
  \label{tabQuanSpinello}%
\end{table*}%

\section{Experiments}

In this section, we evaluate the proposed method using RGBD data collected in a live Intensive Care Unit (ICU) as well as a publicly available pedestrian dataset. We report both quantitative and qualitative results in Section~\ref{secExpQuant} and Section~\ref{secExpQuali}, respectively.

\begin{figure*}
\centering
\includegraphics[width=0.92\linewidth]{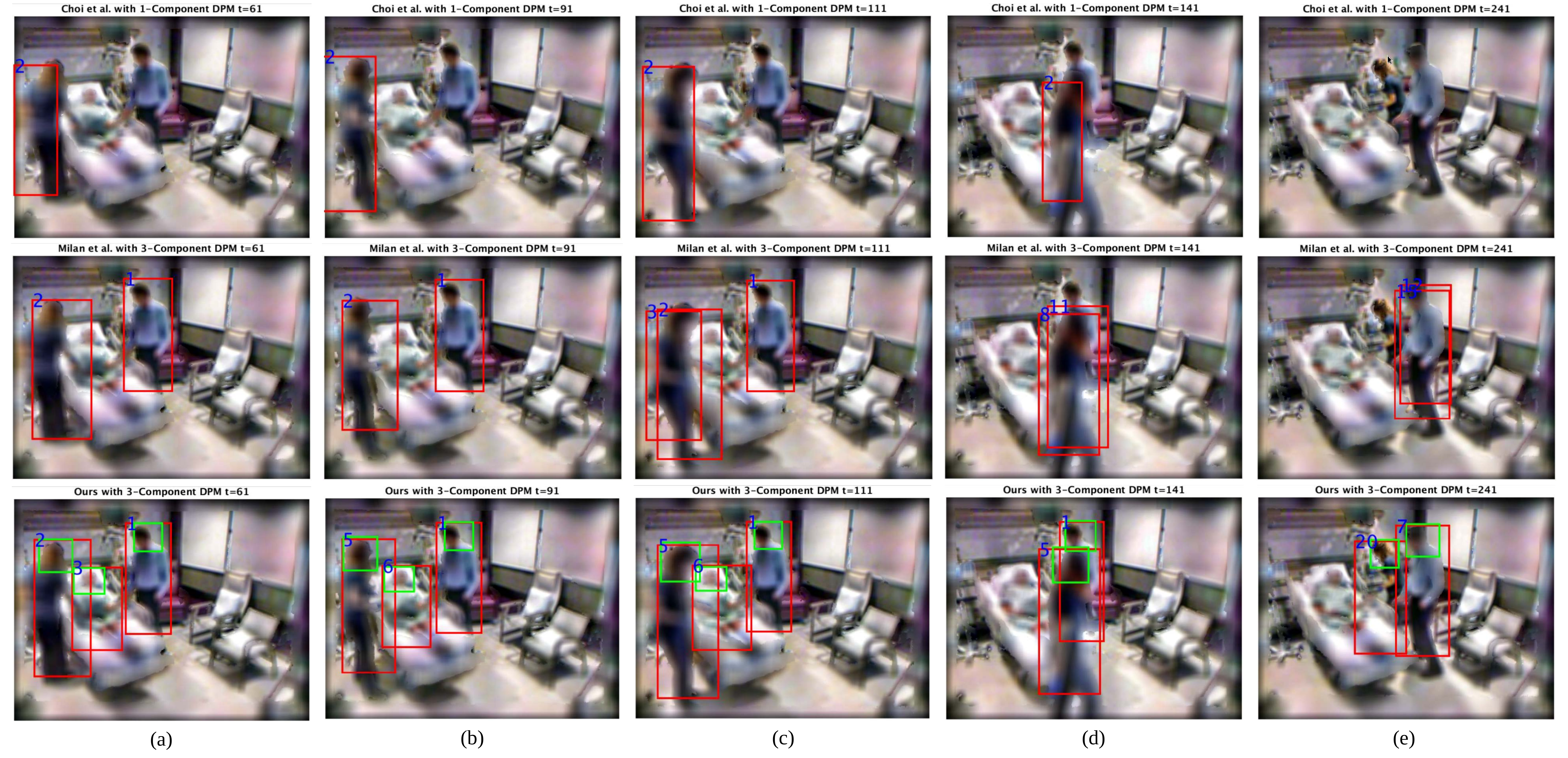}
\caption{Example tracking results on the ICU data, top row: Choi~\etal~\cite{choi2013general}, middle: Milan~\etal~\cite{milan2013continuous}, bottom: ours (better view in color)}
\vspace{-10pt}
\label{figVisualICU}
\end{figure*}


\textbf{Datasets:} In our work, we are primarily motivated by person tracking in complex, structured environments such as an ICU. We report results on data that are collected in a live ICU environment using consumer RGBD cameras at a resolution of 480$\times$640 at an adaptive sampling rate of 30 fps or 1 fps depending on the degree of motion in the room. The dataset contains 36 annotated sequences, each containing 1000 images, capturing 3 different patients. Head and full-body bounding boxes with a sub-sampled rate of 1/10 from the original sequences are annotated\footnote{The detection outputs, ground truth annotations and depth images can be released to public for evaluation.}. The current study is approved by the Institutional Review Board.

We also evaluate our method on a public available dataset. While there are many video datasets \cite{andriluka2010monocular, ferryman2010pets2010, Ferryman2009pets2009, song2013tracking, wu2013online} in the tracking literature, these datasets only provide single object annotations \cite{song2013tracking, wu2013online} or focus on RGB tracking \cite{andriluka2010monocular, ferryman2010pets2010, Ferryman2009pets2009}. For RGBD based multi-target tracking, We use a RGBD pedestrian\footnote{This dataset contains mostly upright walking and standing persons. In spite of lack of significant pose variations as in the ICU data, we still evaluate our method on this dataset, since no other public data can be found.} dataset \cite{spinello2011people} which contains 3 RGBD sequences from three vertically mounted RGBD sensors. There are over 1000 images in each sequence.

\textbf{Experimental Setup:} For preprocessing, background subtraction is simply performed by subtracting the largest depth values over time for each depth image. Detected bounding boxes containing more than 60\% background pixels are removed. HOG features are extracted from both color and depth images. In order to avoid overfitting of the appearance model to any single patient, we use images of 10 sequences from the two patients to train full-body and head detectors by 3-component deformable part models (DPMs)~\cite{felzenszwalb2010object}. For the ICU data, sequences
from the third patient are used for testing. For the pedestrian dataset, we select 1 component (corresponding to standing) in the trained 3-component DPM as the detector, since it contains mostly upright persons. All the 3 sequences are used for testing.

The widely used CLEAR MOT metrics~\cite{keni2008evaluating} are employed for quantitative evaluation. The multiple object tracking precision (MOTP) measures the total error in estimating precise target positions compared with the ground truth. The multiple object tracking accuracy (MOTA) is computed by 100\% minus three types of errors, false positive rate (FP), missed detection rate (MD) and identity switch rate (IDs).

For parameter selection, we empirically normalize the 6 function costs in~\eqref{eqOptSpaTem} to ensure equal impotentness and set the parameters as 1. The number of deformable configurations $C$ is set as 4 for 4 types of pose variations. We assume equal weight for the overlapping ratio and depth information to define the grouping probability in~\eqref{eqProbAll} and set $a$ as 0.5. The probability thresholds $\tau, \delta$ are set as 0.5.

Our method is compared with two state-of-the-art tracking-by-detection algorithms by Choi~\etal~\cite{choi2013general} and Milan~\etal~\cite{milan2013continuous}. We use the publicly available implementations, i.e., Choi~\etal\footnote{http://cvgl.stanford.edu/data2/pr2dataset/} and Milan~\etal\footnote{http://www.milanton.de/contracking/}, and default parameters in their codes, for our experiments\footnote{It should be noticed that we used the RGB version in Choi~\etal~\cite{choi2013general}, since the RGBD version is not available}. Since Choi~\etal~\cite{choi2013general} combine multiple detectors using a fixed spatial configuration in score level, multiple component DPM is not applicable. Thus, 1-component DPM is used for their method on the ICU data. In addition, a pretrained detector is encoded in \cite{choi2013general}. We also report the results using this detector for their method.

\vspace{-2pt}
\subsection{Quantitative Results}\label{secExpQuant}
The averaged overall performance measurements MOTA and MOTP as well as individual measurements on the ICU data are recorded in Table~\ref{tabQuanICU}. From Table~\ref{tabQuanICU}, we can see that our method achieves the highest MOTA and MOTP by combining multiple detectors for different body parts (full-body and head) and modeling the deformable spatial relationship between them. Since person pose changes significantly on the ICU data, the results with an ICU 3-component DPM (multiple components corresponding to multiple poses) are better than those with an ICU 1-component DPM. When using the same full-body detector (i.e., ICU 1-Component DPM), our method outperforms the two state-of-the-art algorithms by MOTA, MOTP, and individual measures except IDs. The MOTA of our method with a 1-component DPM is even higher than that of Milan \etal with a more discriminative 3-component DPM. This shows that our method can outperform existing single-strong-detector-based algorithms by combining multiple weaker detectors.

The results on the public RGBD pedestrian dataset are shown in Table~\ref{tabQuanSpinello}. Though our method cannot achieve significant improvements like those for the ICU data, the MOTA of our method is higher than others and the MOTP is comparable with the highest one by Milan \etal. While these results indicate that single-detector-based algorithm could achieve comparable results for multi-target tracking without extensive pose variations, our method can still achieve good performance by combining multiple detectors under this environment.

\vspace{-2pt}
\subsection{Qualitative Results}\label{secExpQuali}

\begin{figure}
\centering
\includegraphics[width=0.96\linewidth]{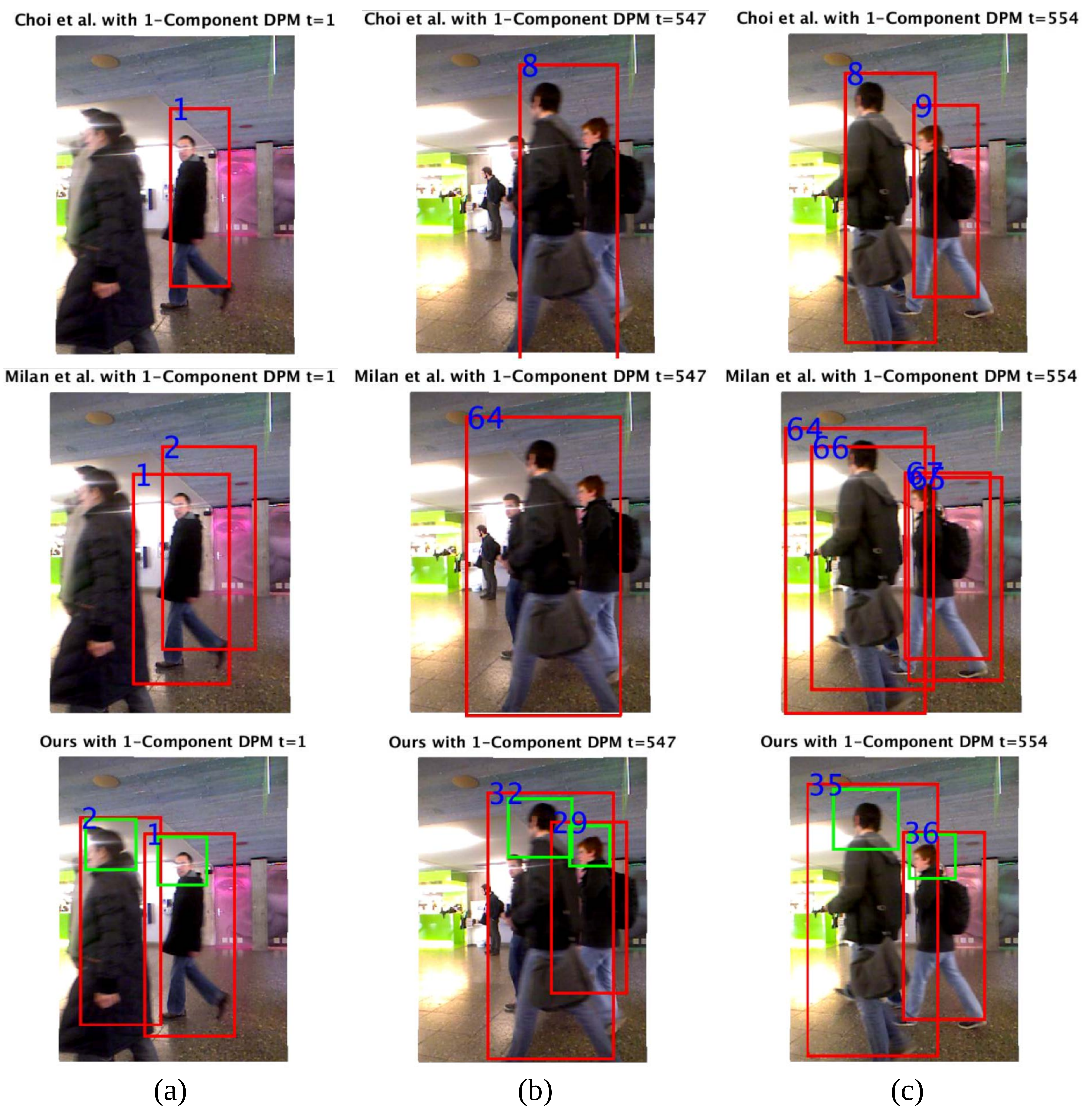}
\caption{Example tracking results on RGBD pedestrian dataset}
\vspace{-10pt}
\label{figVisualSpinello}
\end{figure}

Example tracking results from one sequence out of 26 on the ICU data and out of 3 on the pedestrian dataset are shown in Fig.~\ref{figVisualICU} and Fig.~\ref{figVisualSpinello}, respectively. From Figs.~\ref{figVisualICU}(a)(b)(c), we can see that our method can track all the persons who are standing or lying. This shows that our method is more robust to significant pose variations by modeling the deformable spatial relationship between detectors. On the other hand, as shown in Figs.~\ref{figVisualICU}(d)(e) and Figs.~\ref{figVisualSpinello}(a)(b), our method also works better by combining multiple part-based (head) detector when part of the person is occluded or out of the boundary of the camera.

\section{Conclusion}
In this paper, we propose an approach to combining multiple detectors and model the deformable spatial relationship between them within a tracking-by-detection framework. Detection outputs are grouped to reduce the number of false positives by the location and depth information. The deformable spatial correlations are modeled by minimizing the least-square errors between the expected positions and the projected ones. Person locations are inferred by minimizing an objective function including detection matching, spatial correlation, mutual exclusion, temporal consistence and regularization constraints. Experimental results on data from a live intensive care unit show that the proposed method significantly improves the multi-target tracking performance over state-of-the-art methods.

{\small
\bibliographystyle{ieee}
\bibliography{reference}
}

\balance

\end{document}